# Natural Language Processing with Deep Learning for Medical Adverse Event Detection from Free-Text Medical Narratives: A Case Study of Detecting Total Hip Replacement Dislocation

Alireza Borjali[a,b], Martin Magnéli[a,b,c], David Shin[a], Henrik Malchau[a,d], Orhun K. Muratoglu[a,b], Kartik M. Varadarajan[a,b,*]

[a] Department of Orthopaedic Surgery, Harris Orthopaedics Laboratory, Massachusetts General Hospital, Boston, MA

[b] Department of Orthopaedic Surgery, Harvard Medical School, Boston, MA

[c] Karolinska Institutet, Department of Clinical Sciences, Danderyd Hospital, Stockholm, Sweden

[d] Department of Orthopaedic Surgery, Sahlgrenska University Hospital, Sweden

* Corresponding author

kmangudivaradarajan@mgh.harvard.edu

Abstract

Accurate and timely detection of medical adverse events (AEs) from free-text medical narratives is challenging. Natural language processing (NLP) with deep learning has already shown great potential for analyzing free-text data, but its application for medical AE detection has been limited. In this study we proposed deep learning based NLP (DL-NLP) models for efficient and accurate hip dislocation AE detection following total hip replacement from standard (radiology notes) and non-standard (follow-up telephone notes) free-text medical narratives. We benchmarked these proposed models with a wide variety of traditional machine learning based NLP (ML-NLP) models, and also assessed the accuracy of International Classification of Diseases (ICD) and Current Procedural Terminology (CPT) codes in capturing these hip dislocation AEs in a multi-center orthopaedic registry. All DL-NLP models out-performed all of the ML-NLP models, with a convolutional neural network (CNN) model achieving the best overall performance (Kappa =

0.97 for radiology notes, and Kappa = 1.00 for follow-up telephone notes). On the other hand, the ICD/CPT codes of the patients who sustained a hip dislocation AE were only 75.24% accurate, showing the potential of the proposed model to be used in largescale orthopaedic registries for accurate and efficient hip dislocation AE detection to improve the quality of care and patient outcome.

## 1. Introduction

Accurate and timely detection of medical adverse events (AEs) is critical for improving the quality of care and patient outcomes. The most reliable AE detection method is manual retrospective record review (RRR) of medical data, which are stored within the electronic medical record (EMR) as a combination of structured and free-text medical narrative data[1–3]. Structured EMR data, e.g. International Classification of Diseases (ICD) and Current Procedural Terminology (CPT) codes, are commonly used for large scale research at institutional levels for AE detection[4]. For instance, the Agency for Healthcare Research and Quality (AHRQ) within the United States Department of Health and Human Services uses ICD codes to generate patient safety indicators[1]. Although processing structured EMR data is easy, they have several limitations. First and foremost, the accuracy of structured EMR data is unknown, and more specifically, ICD codes have been shown to be erroneous[5]. For instance, one study tried to evaluate the accuracy of ICD codes for 485 randomly selected patients based on their medical chart, and found that in 30% of medical, and 19% of surgical patients the ICD codes were assigned incorrectly[6]. Another study revealed that only 54% of AEs (out of 2000 patients) following hip arthroplasty surgery had a correct ICD code in a multicenter study[7]. A further limitation of ICD codes is their completeness and level of detail that might not capture all AEs[8,9]. Additionally, since ICD and CPT codes are developed for reimbursement purposes, their appropriateness for medical studies is questionable[5].

An alternative approach to analyzing ICD and CPT codes is to use free-text medical narratives as a data source for AE detection [5]. Medical narratives (e.g. radiology report, discharge summary, operative notes, etc.) are rich source of information. The free format of these narratives enables the clinician to describe the AE with adequate details and contextual information that might not necessarily fall under a specific category within structured codes [10]. Furthermore, medical narratives are written and signed by liable and traceable individuals that can potentially increase their accuracy, as opposed to codes that are susceptible to multiple sources of error (e.g. clinicians, medical coders, upcoding, etc.[9]). The main challenge in analyzing medical narratives is that they are unstructured free-text data. Performing RRR of medical narratives for AE detection is time consuming and requires trained reviewers who are familiar with the report context, structure, and terminology.

Another challenge for accurate AE detection is accounting for incidents that occur outside of hospital. For instance, one study found that as many as 19% of patients (400 patients) experienced some form of AE after discharge from hospital [11]. The main sources of data for detecting these type of AEs are follow-up telephone call records, postal/email questionnaires, and timely follow-up visit notes [12,13]. Analyzing these data is also challenging due to their free-text format. Specifically, follow-up telephone call notes are usually in conversational language, which does not necessarily have a standard structure more commonly seen in other free-text medical narratives; e.g. radiology reports.

The sheer size of medical narratives and their high production rate make accurate and efficient manual analysis impossible. Hence, an automatic workflow for information extraction from free-text medical narratives data in EMR with minimal human interaction is required for accurate and timely AE detection.

Natural language processing (NLP) is a branch of artificial intelligence (AI) concerned with interpreting human language (e.g. written English). NLP can be further divided into three main categories: 1) classical NLP, 2) traditional machine learning based NLP (ML-NLP), and 3) deep learning based NLP (DL-NLP). Classical NLP (also referred to as rule-based NLP) has been extensively applied to medical narratives for information extraction[14]. Classical NLP relies on a variety of manually defined set of rules (e.g. regular expression patterns, terminology lookup, dictionary etc.) for extracting a specific information from free-text data. Defining these set of rules can be challenging, and one set of rules that are defined for a specific database might not be transferable to another. ML-NLP uses labeled data to "learn" a specific task/ classification [15]. These NLP models generally work by tokenizing the free-text into a set of words/regular expressions and then training a classifier (e.g. shallow neural network, support vector machine [SVM], etc.) on them. ML-NLP usually achieves higher performance compared with classical NLP [16]. Some researchers have also used a hybrid approach by combining ML-NLP and classical NLP, usually defining a set of heuristic rules to handle ML-NLP errors [15].

DL-NLP is an emerging category that has already produced impressive results in many domains [17]. DL-NLP methods consist of multiple processing layers including/ inspired by neural networks, which get trained on the free-text data without the need for defining any hand-crafted rules. It has been shown in the literature that even a simple DL-NLP can outperform classical and ML-NLP methods [17]. Two of the most frequently used DL-NLP methods are recurrent neural networks (RNNs) and convolutional neural networks (CNNs), which have been successfully applied in different domains for various free-text analysis tasks such as text classification [18], sentiment analysis [19], summarization[20], and machine translation [21].

Despite the overwhelming success of DL-NLP methods in many domains, their application for AE detection from medical narratives is lacking. A recent (December, 2019) systematic review of NLP studies for AE analysis found no DL-NLP study in the literature [10]. Another recent (March, 2020) survey only found a handful of studies solely focusing on adverse drug event (ADE) detection [22]. Furthermore, the current literature on using NLP for AE detection has mainly focused on analyzing standard data inputs such as radiology reports, discharge summaries, and incident reports [10,22]. These types of data usually follow a standard structure and are written with a specific technical terminology. On the other hand, other data sources such as notes of follow-up telephone calls are in conversational English. Patients and providers talk differently in conversation and may describe the same AE in various ways—notes of such conversations may be more difficult to analyze compared with standard medical narrative free-text data. We hypothesized that state-of-the-art DL-NLP methods can be used for accurate and efficient AE detection from both standard (e.g. radiology reports) and non-standard (e.g. follow-up telephone calls notes) free-text medical narratives. Hence, the objectives of this study were as follows:

1- To evaluate the potential of leveraging DL-NLP methods for accurate and efficient in- and out-of-hospital AE detection from different free-text medical narratives (i.e., standard radiology reports and non-standard conversational follow-up telephone call notes.)
2- To compare the performance of DL-NLP methods with the established ML-NLP methods for AE detection from free-text medical narratives.
3- To measure the accuracy of ICD/CPT codes for capturing AE and compare that with the accuracy of DL-NLP methods AE detection.

To achieve these objectives, we selected hip dislocation after primary total hip replacement (THR) as the AE of interest as a case study. Hip dislocation is one of the most common AEs following THR surgery and the reported dislocation rate in previous studies ranges from 0.2 to 10% [23–25]. Dislocation is also one of the major reasons for revision surgery, reported to be the etiology for 11 to 24% of revision cases [26,27]. In the major arthroplasty registries, the rate of dislocation resulting in revision surgery is presented, however the overall dislocation rate is not [26,27]. Furthermore, the accuracy of arthroplasty registries in capturing hip dislocation events that happen out-of-hospital is unknown, since these registries are typically developed based on data from direct patient encounters with hospitals and reliant on manual or code-based reporting of AEs in such encounters. We hypothesized that the accuracy of hip dislocation AE detection in an arthroplasty registry can be improved by leveraging DL-NLP methods to analyze both in-hospital (radiology reports) and out-of-hospital (telephone call notes) free-text medical narratives.

## 2. Materials and methods

### 2.1 Data

The study cohort consisted of all primary THR patients operated within a regional network of multiple hospitals between January 2016 and June 2019. Institutional review board (IRB) approval was acquired prior to collecting the data. All patients were registered in the institutional arthroplasty registry of the hospital network for primary THR. We assumed the patients who suffered a hip dislocation AE post primary THR surgery either came back and were treated within the hospital network, or were treated outside the hospital network.

We hypothesized that the patients who sought treatment for hip dislocation AE within the hospital network would have radiographs taken of their hip or pelvis. These images were recorded within an Epic-based EMR system, and were accompanied by free-text narrative of the radiologist's assessment. In particular, the 'Impression' field of the assessment contained the radiologist's summary description of the main findings. The language used within these summary descriptions was found to be concise and uniform. All such available radiologist summary descriptions—hereafter described as "radiology notes"—were retrieved for all postoperative images of the hip or pelvis within the study period extended by 90 days.

We further hypothesized that patients who were treated for hip dislocation AE outside of the hospital network might not have such radiologist assessments available. In order to capture such "external" dislocation events, free-text fields in the EMR describing follow-up telephonic conversations with the patients—hereafter described as "telephone notes"—were retrieved for all study patients (if such notes were available) within the study period extended by 90 days.

There were 6617 primary THR patients in the registry with 7156 surgeries (some patients had both hips replaced). A preliminary filter was applied to the radiology and telephone notes of all patients to exclude any notes without any of the following strings: "disloc," "sublux," "reduc," "reloc," "displace." The aim of this filter was to reduce the number of irrelevant notes to be assessed without excluding notes describing real incidences of hip dislocation AE. After this initial filtering, 1890 patients were selected for detail record review resulting in 3014 radiology and 783 telephone notes.

A two-member review team including a senior resident orthopaedic surgeon and a registry staff member separately reviewed all radiology (3014) and telephone (783) notes. Inter-rater agreement was determined following the independent assessments and any discrepancies were resolved by consensus. In total, 105 (out of 1890) patients sustained a dislocation. This generated a total of 380 radiology and 174 telephone notes indicating a dislocation, and 2634 radiology and 609 telephone notes with no indication of a dislocation.

The radiology notes were labeled into three categories as follows:

1. **Current dislocation:** Radiology note that indicated a current dislocation e.g. "the radiograph shows a dislocated hip arthroplasty" (247 radiology notes).

2. **Evidence of previous dislocation:** Radiology note that indicated that the patient had sustained a dislocation and had been reduced e.g. "radiograph shows a successful reduction of previous dislocated hip arthroplasty" (133 radiology notes).

3. **No dislocation:** Radiology note that did not indicate that the patient had sustained a dislocation, e.g. "the radiograph shows a total hip arthroplasty without signs of fracture or dislocation" (2634 radiology notes).

The telephone notes were labeled into two categories as follows:

1. **Evidence of previous dislocation:** Telephone note that indicated that the patient had sustained a dislocation and had been reduced e.g. "Patient was seen in the ED at (name is deleted) Hospital with right hip dislocation on (date is deleted). He underwent a right hip closed reduction under anesthesia …" (174 telephone notes).

2. **No dislocation:** Telephone note that did not indicate that the patient had sustained a dislocation e.g. "Pt was asked to inform GZ of her visit to the (name is deleted) ER on (date is deleted), pt had a hip replacement completed last month and the concern was a possible hip dislocation. Pt would like to report this was not the case, that she had suffered from a "very badly pulled muscle". Pt can be contacted with any additional questions ..."   (609 telephone notes).

It is important to mention that the telephone notes were collected at specific follow-up time points; hence they did not capture any "current dislocation" as opposed to the radiology notes.

Furthermore, each record was treated individually as a data point, i.e., the NLP was designed to analyze each record individually and decide if that record alone had any indication of hip dislocation AE.

After identifying all patients with "current dislocation" and "evidence of previous dislocation", either from radiology notes or telephone notes, we identified ICD-10 and CPT codes recorded within the EMR for these patients. These codes are described in Table 1. We compared the accuracy of ICD-10 and CPT codes in capturing a hip dislocation AE with different NLP methods using manual record review as the gold standard.

**Table 1** Current Procedural Terminology Code (CPT) and International Classification of Diseases, 10$^{th}$ Revision (ICD-10) codes indicating post hip arthroplasty dislocation adverse event

| Type | Code | Description |
| --- | --- | --- |
| *CPT | 27265 | Closed treatment of post hip arthroplasty dislocation; without anesthesia |
| *CPT | 27266 | Closed treatment of post hip arthroplasty dislocation; requiring regional or general anesthesia |
| **ICD-10 | T84.020A | Dislocation of internal right hip prosthesis, initial encounter |
| **ICD-10 | T84.020D | Dislocation of internal right hip prosthesis, subsequent encounter |
| **ICD-10 | T84.020S | Dislocation of internal right hip prosthesis, sequela |
| **ICD-10 | T84.021A | Dislocation of internal left hip prosthesis, initial encounter |
| **ICD-10 | T84.021D | Dislocation of internal left hip prosthesis, subsequent encounter |
| **ICD-10 | T84.021S | Dislocation of internal left hip prosthesis, sequela |

*CPT: Current Procedural Terminology Code
**ICD-10: International Classification of Diseases, 10$^{th}$ Revision

## 2.2 Experimental setup

We performed two sets of experiments: one on the radiology notes and the other one on the telephone notes.

*2.2.1 Radiology notes*

We developed different ML- and DL-NLP models to categorize the radiology notes into three categories: 1) current dislocation, 2) evidence of previous dislocation, and 3) no dislocation. We compared the models' performance against each other. We also compared the accuracy of the best performing model against the accuracy of ICD-10 and CPT codes in capturing hip dislocation AE.

*2.2.1 Telephone notes*

We developed different ML- and DL-NLP models to categorize the telephone notes into two categories: 1) evidence of previous dislocation, and 2) no dislocation. We followed the same approach as section 2.2.1 and compared the models' performance against each other. We also compared the accuracy of the best performing model against the accuracy of ICD-10 and CPT codes in capturing hip dislocation AE.

**2.3 Proposed models**

We implemented multiple DL-NLP models and benchmarked their performance against ML-NLP models. We did not assess classical rule-based NLP models, since their performance has already been compared with ML-NLP models, with the general consensus being that the latter outperforms the former [16]. The implemented models are explained in the following sections.

*2.3.1 Traditional machine learning based natural language processing (ML-NLP) models*

The ML-NLP models implemented in this study included text preprocessing and tokenization (similar among all models), and final classification (different classifier for each model). During text preprocessing, any non-letter characters (e.g. special characters, digits, etc.) were removed from the text. The text was normalized by removing word suffixes using Porter stemming algorithm [28]. Then, the text was tokenized using linguistic tokens to break the text into sequence of words. These tokens were then fed into the classifier using unigram or n-gram (n= 2, 3) language models. Different classifiers including Generalized Linear Regression, K-NN, Random Forest, SVM, and Shallow Neural Network were implemented. Detailed explanations of these models can be found elsewhere [29–34].

*2.3.2 Deep learning based natural language processing (DL-NLP) models*

Two DL-NLP models were developed: 1) multilayer bidirectional long short-term memory (LSTM) RNN, and 2) CNN. Both models were implemented using Tensorflow (Keras) on a workstation comprised of an Intel(R) Xeon(R) Gold 6128 processor, 64GB of DDR4 RAM, and an NVIDIA Quadro P5000 graphic card. We explain each model's architecture and design rationale in the following sections. While discussion of the mathematical details of these models is out of the scope of this study, the references cited offer further explanation.

*2.3.2.1 Proposed long short-term memory (LSTM) model*

RNN is a class of neural networks designed for handling sequential data (e.g., free-text data). In an RNN model, the output of each step is a function of the output of the previous steps. In other words, an RNN model has the "memory" to look back at what has been calculated so far at each step, and to incorporate that into its calculation for the current step. In practice, however, RNN models are limited to looking back only by a few steps due to the vanishing (or exploding) gradient problem [35]. LSTM is a variation of vanilla RNN that remedies this issue [36]. LSTM uses a mechanism called "cell state" to carry information along different steps. The cell state gets updated through regulated structures called "gates". There are three main gates in an LSTM model: "input gate", "forget gate", and "output gate". These gates determine what new information should get through to update the cell state (input gate), what information should be discarded (forget gate), and finally, what information should be output based on the cell state (output gate). This structure enables LSTM to capture long-range dependencies in a sequential data (e.g., free-text) [37].

Bidirectional LSTM (BiLSTM) is an improvement of vanilla LSTM that considers both preceding and succeeding long-range dependencies of the current step. More specifically, BiLSTM for NLP application "looks at" the words before and after a given word while analyzing each word in a text. BiLSTM uses two LSTMs: one gets trained on the input sequence as is, and the other one gets trained on the reversed input sequence to capture dependencies in both directions. Another variation of LSTM is a multilayer LSTM. Multilayer LSTM stacks layers of LSTMs on top of each other to create a deeper model and increase the overall computation capacity for more complicated tasks [38].

We implemented a multilayer BiLSTM RNN with a word embedding input layer. Word embedding maps each word into a real-valued vector. The vectors are initialized with random numbers that will get updated during the training process based on the usage of the words in the text [39]. As a result, words that have similar meaning will have similar vector representation. We used Word2vec algorithm [40] with 1000-dimension (1000-D) vector space.

The model had two BiLSTM layers. Although identifying the exact contribution of each layer is not straightforward, we chose two layers to analyze a given text at different abstraction levels, i.e., word, and sentence level. Inputs to the first layer were the words of the text, while the input to the second layer was the output from the first layer, which depended on each word and what came before and after it; hence, the second layer had sentence level inputs. We chose BiLSTM layers to consider the dependencies in both directions. For instance, considering these sentences: "No fracture or dislocation was observed" vs. "Dislocation was previously reduced". In the first sentence, what precedes the word "dislocation" defines the status of dislocation, while in the second sentence what succeeds the word "dislocation" defines its status (no dislocation vs. evidence of previous dislocation). We used a fully-connected neural network classifier. All hyper-parameters (e.g. LSTM hidden layers, fully-connected layers, training parameters, etc.) where optimized using the validation dataset (Table 2).

*2.3.2.2 Proposed convolutional neural network (CNN) model*

CNN models have been primarily used for computer vision tasks [41–45] with growing application in NLP [46,47]. CNNs can extract features from a given text by convolving a filter over it to create features that resemble n-grams (n is equivalent to the size of the filter). Then these features are fed into a classifier to get a final classification for a specific task [17].

We implemented a CNN model with two 1-D convolution layers following the same design rationale as the proposed LSTM model. The first layer's filter size was 3 to create 3-gram like features on word level. The second convolution layer's filter size was 5 to combine the previous layer's output and look at the features on sentence level. We also used a fully-connected neural network classifier. All hyper-parameters (e.g. CNN hidden layers, fully-connected layers, training parameters, etc.) where optimized using the validation dataset (Table 2).

**Table 2** Details of the proposed deep learning based natural language processing models

| Proposed *LSTM | | Proposed **CNN | |
|---|---|---|---|
| Layer | Description | Layer | Description |
| Input | Radiology/ Telephone Notes | Input | Radiology/ Telephone Notes |
| Word Embedding Layer | Vector Dimension = 1000, Maximum Number of Words = 3000, Maximum Sequence Length = 256 | Word Embedding Layer | Vector Dimension = 1000, Maximum Number of Words = 3000, Maximum Sequence Length = 256 |
| 1st Bidirectional LSTM | Hidden Layers = 64 | 1st Convolution Layer | 1-D Filter size = 3, Stride = 1, Hidden Layers = 128, Activation= ReLU |
| 2nd Bidirectional LSTM | Hidden Layers = 64, Dropout = 0.2, Recurrent Dropout = 0.4 | 2nd Convolution Layer | 1-D Filter size = 5, Stride = 1, Hidden Layers = 128, Activation = ReLU |
| Classifier 1st Dense Layer | Hidden Layers = 64 for Radiology Notes/ 128 for Telephone Notes, Activation = ReLU | Classifier 1st Dense Layer | Hidden Layers = 64, Activation = ReLU |
| Output (Classifier 2nd Dense Layer) | Hidden Layers = 3, Activation = SoftMax for Radiology Notes/ Hidden Layers = 1 , Activation = Sigmoid for Telephone Notes | Output (Classifier 2nd Dense Layer) | Hidden Layers = 3, Activation = SoftMax for Radiology Notes/ Hidden Layers = 1 , Activation = Sigmoid for Telephone Notes |

\* Long Short Term Memory          \*\*Convolutional Neural Network

## 2.4 Model evaluation

We used a split validation method [48] and divided both datasets (radiology and telephone notes) into training, validation, and test subsets with an 80:10:10 split ratio. All models were trained on the training subset and the validation subset was used for tuning the hyper-parameters. Ultimately, the trained models were tested on the test subset, which was isolated from the training and validation process, and the outcomes were reported as the models' performance. The performance of the best overall model was compared with the accuracy of ICD-10/CPT codes in identifying the same hip dislocation AEs.

# 3. Results

## 3.1 Radiology notes

Table 3 shows the results of both ML- and DL-NLP models classifying the radiology notes in the test subset (301 notes; held out during model training) into three categories as follows: 1) no dislocation (263 notes), 2) current dislocation (25 notes), and 3) evidence of previous dislocation (13 notes).

**Table 3** Results of different natural language processing (NLP) models for detecting hip dislocation adverse event (AE) from radiology notes

| Model | | | No dislocation | | Current dislocation | | Evidence of previous dislocation | | Kappa |
|---|---|---|---|---|---|---|---|---|---|
| | | | Class recall [%] | Class precision [%] | Class recall [%] | Class precision [%] | Class recall [%] | Class precision [%] | |
| Traditional machine learning | Generalized Linear Model | Unigram | 98.5 | 94.2 | 40.0 | 52.6 | 46.1 | 85.7 | 0.53 |
| | | 2-gram | 98.1 | 94.5 | 48.0 | 63.2 | 58.3 | 87.5 | 0.61 |
| | | 3-gram | 98.1 | 95.2 | 56.0 | 66.7 | 50.0 | 83.3 | 0.63 |
| | K-NN | Unigram | 94.3 | 93.9 | 52.0 | 41.9 | 46.1 | 100.0 | 0.50 |
| | | 2-gram | 92.0 | 95.3 | 64.0 | 38.1 | 23.1 | 60.0 | 0.47 |
| | | 3-gram | 92.0 | 94.5 | 60.0 | 39.5 | 38.5 | 71.4 | 0.47 |
| | Random Forest | Unigram | 98.5 | 92.2 | 32.0 | 40.0 | 0.0 | 0.0 | 0.37 |
| | | 2-gram | 99.6 | 88.2 | 12.0 | 75.0 | 0.0 | 0.0 | 0.12 |
| | | 3-gram | 98.5 | 88.7 | 8.0 | 22.2 | 0.0 | 0.0 | 0.11 |
| | SVM | Unigram | 79.4 | 97.1 | 35.3 | 12.0 | 45.4 | 20.0 | 0.24 |
| | | 2-gram | 89.3 | 95.9 | 48.0 | 24.5 | 7.7 | 14.3 | 0.35 |
| | | 3-gram | 92.0 | 94.2 | 56.0 | 31.8 | 0.0 | 0.0 | 0.38 |
| | Shallow Neural Network | Unigram | 100.0 | 87.4 | 0.0 | 0.0 | 0.0 | 0.0 | 0.00 |
| | | 2-gram | 100.0 | 87.4 | 0.0 | 0.0 | 0.0 | 0.0 | 0.00 |
| | | 3-gram | 100.0 | 87.4 | 0.0 | 0.0 | 0.0 | 0.0 | 0.00 |
| Deep learning | Proposed LSTM | | 99.2 | 98.5 | 84.0 | 87.5 | 92.3 | 100.0 | 0.89 |
| | Proposed CNN | | 100.0 | 99.6 | 96.0 | 96.0 | 92.3 | 100.0 | 0.97 |

Both DL-NLP models (proposed CNN and LSTM) outperformed all ML-NLP models and achieved the highest and second highest Kappa score. The CNN model achieved the highest overall results for all three classes.

ML-NLP models generally performed well in classifying "no dislocation" notes. However, they struggled in classifying the "no dislocation" notes that mentioned some other form of dislocation, e.g. dislocated fracture fragments, and/or, notes where the status of dislocation was defined by long range dependencies. e.g. "no dislocation" where "no" came immediately before dislocation vs. "no fracture or dislocation" where to understand the status of dislocation the model needed to look further back in the sentence and relate "no" to "dislocation." Figure 1 shows an example of such notes that were misclassified by the ML-NLP models. On the other hand, both DL-NLP models classified these type of notes correctly, where the CNN model achieved 100% class recall and the LSTM model only made a few misclassifications where the dislocation was described in an atypical fashion—including uncommon terms such as "dislocated *hemiprosthesis*" and "*posterosuperior* dislocation", which only appeared in one and two notes respectively in the entire dataset.

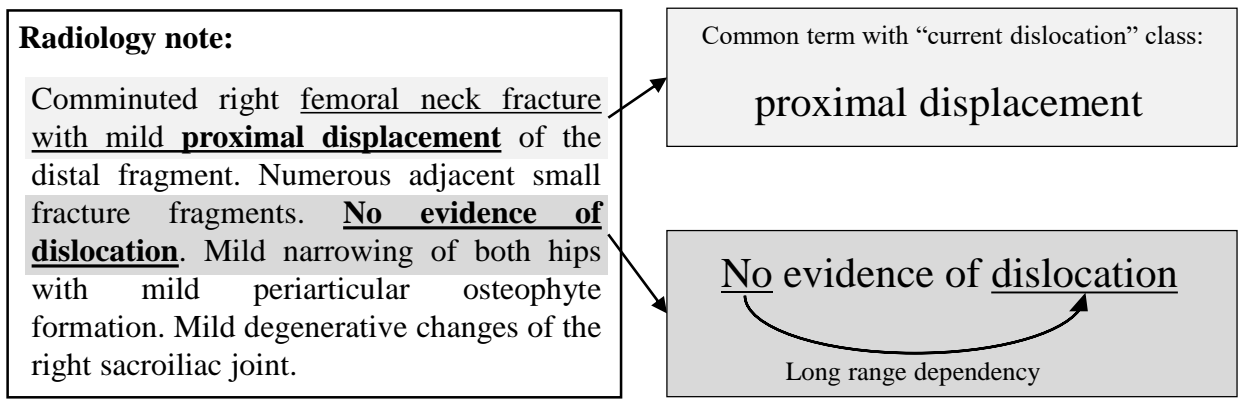

**Figure 1** An example of "no dislocation" radiology note that was misclassified as "current dislocation" by traditional machine learning based natural language processing (ML-NLP) models.

ML-NLP models did not generally achieve high performance classifying either "current dislocation" or "evidence of previous dislocation" radiology notes. For instance, with the following radiology note: "Reduction of previous dislocation of the left total hip prosthesis. No dislocation or fracture is seen now." These models could not distinguish between "no dislocation" and "evidence of previous dislocation" and misclassified this note as "no dislocation."

On the other hand, DL-NLP models achieved high performance classifying "current dislocation" and "evidence of previous dislocation" radiology notes. The CNN model only made two misclassifications overall (299 correct out of 301 radiology notes in the test subset). These two notes were unique in the entire dataset and did not only focus exclusively on the hip. Figure 2 shows an example radiology note misclassified by the CNN (and LSTM) model. This note discussed dislocation and displacement at multiple anatomic sites. Although the hip was dislocated, the knee was not, and there was no displaced rib; hence the proposed CNN misclassified this radiology note as "no dislocation".

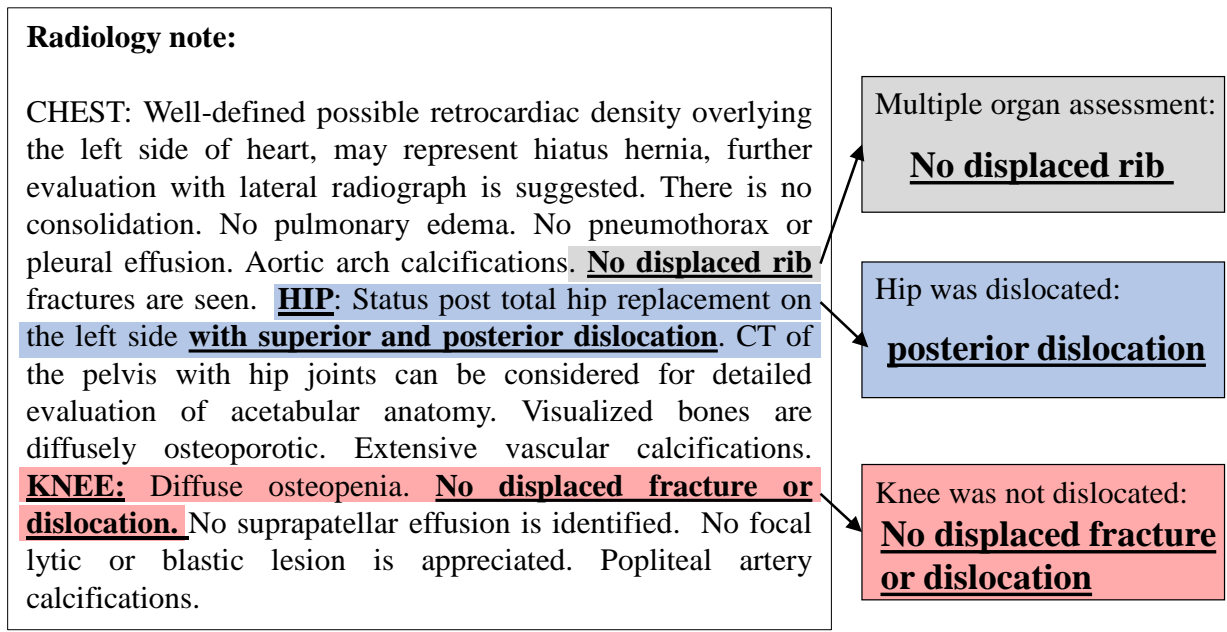

**Figure 2** An example of "current dislocation" radiology note that was misclassified as "no dislocation" by deep learning models.

*3.2 Telephone notes*

Table 4 shows the results of both ML- and DL-NLP models classifying the telephone notes in the test subset (held out from models during training) into two categories as follows: 1) evidence of previous dislocation (17 notes), and 2) no dislocation (62 notes).

**Table 4** Results of different natural language processing (NLP) models for detecting hip dislocation adverse event (AE) from telephone notes

| Model | | | No dislocation | | Evidence of previous dislocation | | Kappa |
|---|---|---|---|---|---|---|---|
| | | | Class recall [%] | Class precision [%] | Class recall [%] | Class precision [%] | |
| Machine learning | Generalized Linear Model | Unigram | 77.8 | 83.3 | 56.2 | 47.4 | 0.32 |
| | | 2-gram | 93.3 | 76.4 | 18.7 | 50.0 | 0.15 |
| | | 3-gram | 88.9 | 75.5 | 7.1 | 16.7 | -0.05 |
| | K-NN | Unigram | 88.9 | 87.0 | 64.7 | 68.7 | 0.55 |
| | | 2-gram | 93.3 | 84.0 | 53.0 | 75.0 | 0.51 |
| | | 3-gram | 95.6 | 86.0 | 58.8 | 83.3 | 0.60 |
| | Random Forest | Unigram | 55.6 | 86.2 | 76.5 | 39.4 | 0.25 |
| | | 2-gram | 68.9 | 83.8 | 64.7 | 44.0 | 0.29 |
| | | 3-gram | 44.44 | 74.0 | 58.8 | 28.6 | 0.02 |
| | SVM | Unigram | 71.1 | 80.0 | 53.0 | 40.9 | 0.22 |
| | | 2-gram | 100.0 | 76.3 | 17.6 | 100.0 | 0.24 |
| | | 3-gram | 100.0 | 73.8 | 5.9 | 100.0 | 0.08 |
| | Shallow Neural Network | Unigram | 100.0 | 72.6 | 0.0 | 0.0 | 0.00 |
| | | 2-gram | 100.0 | 72.6 | 0.0 | 0.0 | 0.00 |
| | | 3-gram | 100.0 | 72.6 | 0.0 | 0.0 | 0.00 |
| Deep learning | Proposed LSTM | | 93.6 | 86.8 | 58.8 | 83.3 | 0.63 |
| | Proposed CNN | | 100.0 | 100.0 | 100.0 | 100.0 | 1.00 |

Both DL-NLP models outperformed all ML-NLP models and achieved the highest and second highest Kappa score. The CNN model achieved the highest overall results.

ML-NLP models generally performed poorly (best Kappa = 0.60). They could not identify the "evidence of previous dislocation" in the telephone notes. Telephone notes were more challenging compared to the radiology notes to extract information from. Figure 3 shows an example of each telephone note category. These notes required more contextual interpretation and varied significantly for each patient, with no standard structure as was commonly seen with the radiology notes. On the other hand, both DL-NLP models achieved higher Kappa values compared with ML-NLP models. The CNN model achieved perfect performance classifying all telephone notes correctly.

| Part of the telephone note | Classification |
|---|---|
| …wasn't sure what actually popped as she was trying to get something out of the fridge yesterday but is moving around OK today. Will check in with her home therapist about best ways to do what she needs to do. She is really being quite cautious and was reassured that if she is moving around comfortably she is **not dislocated**. | No dislocation |
| ... she was sitting with her legs bent and was putting them back down when she heard a crack and increase pain. She did not go to the ER until later, they did an x-ray and it showed a **dislocation**. They gave her sedation and were able to reduce the dislocation. She was just discharged home this morning. | Evidence of previous dislocation |

**Figure 3** Example of "no dislocation" and "evidence of previous dislocation" telephone notes.

*3.3 Current Procedural Terminology (CPT) and International Classification of Diseases, 10$^{th}$ Revision (ICD-10) codes error analysis*

Among 105 patients who sustained a prosthetic hip dislocation AE, 26 patients did not have any relevant coding (Table 1) in their EMR. CPT/ ICD-10 coding only captured 75.24% of all hip dislocation AEs. Among these 26 patients, 16 patients had dislocations that could be identified only by the telephone notes, 7 patients only by radiology notes, and 3 patients by both.

## 4. Discussion

In this study we proposed two DL-NLP models for efficient and accurate hip dislocation AE detection from standard (radiology notes) and non-standard (telephone note) free-text medical narratives. We also benchmarked these proposed DL-NLP models performance with a wide variety of ML-NLP models, and the accuracy of ICD-10 and CPT codes in capturing hip dislocation AE.

Few other studies [50–53] have used DL-NLP models for AE detection from free-text medical narratives solely focusing on ADE. Dev et al. used single layer LSTM model for binary classification of adverse drug events as serious vs. non-serious [52]. They also benchmarked their DL-NLP with ML-NLP models and reported higher performance for DL-NLP. However, they did not benchmark their DL-NLP with other more advanced models (e.g., multilayer BiLSTM, CNN, etc.). They also only used standard medical narratives, e.g. AE reports. Dandala et al. and Xu et al. used BiLSTM based models for classification of ADEs [51,53]. However, they did not benchmark their DL-NLP models against ML-NLP models, and only used standard medical narratives (clinical notes). Huynh et al. implemented different DL-NLP models for binary classification of ADE [50]. They used two source of narratives (non-medical from Twitter dataset) and standard medical (ADE case reports), and benchmarked their DL-NLP models with ML-NLP models. They reported higher performance for all the DL-NLP models compared with the ML-NLP models, where CNN model achieved the best overall performance. While other studies have applied DL-NLP models on medical free-text narratives for ADE binary classification, to the best of our knowledge, this is the first study to apply such models to automatically detect hip dislocation AE from standard and non-standard medical free-text narratives, and perform binary and categorical classifications.

The CNN model proposed in this study achieved the best overall performance (Kappa = 0.97) in classifying the radiology notes into three categories: 1) no dislocation, 2) current dislocation, and 3) evidence of previous dislocation), as well as best overall performance (Kappa = 1.00) classifying the telephone notes into two categories: 1) no dislocation, and 2) evidence of previous dislocation. This proves our hypothesis that state-of-the-art DL-NLP model can be used for accurate and efficient AE detection from free-text medical narratives.

ML-NLP models generally performed better in analyzing the radiology notes (best Kappa = 0.63) compared with the telephone notes (best Kappa = 0.24). This showed that different medical narratives pose different level of challenges with regards to application of NLP models. Radiology notes are written in a standard format with more uniform terminology, while telephone notes are more conversational in language and format and variable between patients. Nevertheless, our proposed CNN model achieved perfect performance for classifying the telephone notes, and only misclassified 2 (out of 301) unique radiology notes where multiple anatomic sites of dislocation/displacement were discussed.

We also investigated the ICD-10/CPT codes of the patients who were identified by free-text notes as sustaining a prosthetic hip dislocation AE. Only 75.24% of these patients had a relevant code in their EMR. We also showed that the majority (61.54%) of these patients who did not have relevant coding in the EMR could only be identified by analysis of their follow up telephone notes.

One limitation of this study was that the models were trained on data from only one hospital network. Although this network consisted of multiple regional hospitals, to what extent these models can be generalized across other hospitals remains unknown. Furthermore, this study only focused on detecting one AE. Although accurate and efficient hip dislocation AE detection is of great value in orthopaedics, application of these models to detect other types of AE requires further investigation. Another limitation was the size of the dataset. Specifically, radiology notes had a very uniform language, and therefore, classifying the outlier cases that did not follow the similar language requires more data. With a larger dataset, DL-NLP model could learn to classify more challenging notes and distinguish, for example, between knee and hip dislocation. One method of improving NLP accuracy in classifying radiology notes without altering models is if radiologists could agree on common terms that should be used to describe certain diagnoses in their department. This could also make it easier for clinicians reading these notes.

In this study, we developed an efficient and accurate DL-NLP model to automatically detect prosthetic hip dislocation AE from standard and non-standard free-text medical narratives. The NLP model in this study was developed on data abstracted from the most frequently used EMR system in the U.S., Epic, and it could potentially be implemented in all orthopaedic departments using Epic-based EMR system. The rate of dislocations could be seen as a quality measure, and an NLP model for dislocations could be used both on a hospital level to compare dislocation rates between different hospitals and on a surgeon level to compare rates of AEs between surgeons. Of course such comparison should be done with caution, adjusting for patient risk factors and not comparing revision cases with primary cases. Before this DL-NLP model can be used on data from other hospitals, we suggest that it get validated on data from such hospitals, following a similar form of RRR as was performed in this study. This way, the true sensitivity and specificity of both the NLP model and ICD and CPT coding could be calculated for different hospitals following the similar approach to this study.